\documentclass[runningheads]{llncs}
\usepackage[T1]{fontenc}
\usepackage{graphicx}
\usepackage{booktabs}
\usepackage[misc]{ifsym}
\newcommand{\corr}{(\Letter)}
\usepackage{mwe}

\usepackage{tikz}
\usetikzlibrary{positioning}
\usepackage{subcaption}
\usepackage{todonotes}
\usepackage[unicode]{hyperref}
\usepackage{algorithm}
\usepackage{algpseudocode}
\usepackage{pdfpages}
\usepackage{amsmath}
\usepackage{amsfonts}
\usepackage{makecell}

\begin{document} 

\title{RangeAD: Fast On-Model Anomaly Detection}


\author{Author information scrubbed for double-blind reviewing}
\author{Luca Hinkamp\inst{1} \corr \orcidID{0009-0004-1547-1590} \and
Simon Klüttermann\inst{1} \orcidID{0000-0001-9698-4339}  \and
Emmanuel Müller\inst{1,2}\orcidID{0000-0002-5409-6875}}

\authorrunning{L. Hinkamp et al.}

\institute{TU Dortmund University, Dortmund, Germany \email{\{luca.hinkamp, simon.kluettermann, emmanuel.mueller\}@cs.tu-dortmund.de}
\and
Research Center Trustworthy Data Science and Security, UA Ruhr, Germany}

\maketitle              

\begin{abstract}
In practice, machine learning methods commonly require anomaly detection (AD) to filter inputs or detect distributional shifts. Typically, this is implemented by running a separate AD model alongside the primary model. However, this separation ignores the fact that the primary model already encodes substantial information about the target distribution.
In this paper, we introduce On-Model AD, a setting for anomaly detection that explicitly leverages access to a related machine learning model. Within this setting, we propose RangeAD, an algorithm that utilizes neuron-wise output ranges derived from the primary model.
RangeAD achieves superior performance even on high-dimensional tasks while incurring substantially lower inference costs. Our results demonstrate the potential of the On-Model AD setting as a practical framework for efficient anomaly detection.
\end{abstract}

\section{Introduction}

Machine learning (ML) has become a cornerstone of modern software systems; however, the transition from controlled experimental settings to practical, real-world deployment requires robust safeguards. Anomaly Detection (AD) is critical to this transition, acting as a gatekeeper for system reliability~\cite{survey-ruff,pm4}. The necessity of AD in production environments is driven by three primary challenges. First, the principle of "garbage in, garbage out" dictates that models perform unpredictably when fed corrupted or irrelevant data; AD serves as a crucial filter to reject such inputs before they degrade system performance~\cite{garbageInGarbageOut}. Second, real-world environments are non-stationary. As data distributions evolve (a phenomenon known as concept drift), AD allows systems to detect when current inputs no longer match the training distribution, signaling the need for adaptation~\cite{ConceptDriftBin}. Finally, standard ML models are typically optimized for the average case and suffer from underspecification when facing "black swan" events or outliers. AD acts as a safety mechanism to identify and handle these edge cases where model guarantees fail~\cite{blackSwanUnderspecificationMl}.

Despite these necessities, integrating AD into production pipelines remains challenging. Typically, AD is deployed as a secondary system running alongside the primary predictive model. This duality increases computational overhead and introduces a risk of misalignment, where the AD model’s definition of "normal" does not perfectly map to the primary model’s operational domain. Furthermore, traditional AD methods are often hindered by the curse of dimensionality~\cite{curseOdim}, which limits the complexity of inputs they can handle effectively. Perhaps most critically, the fundamental unknowability of future anomalies makes hyperparameter tuning and model selection a near-impossible task; optimizing for known anomalies often fails to generalize to unforeseen contingencies encountered in the wild~\cite{hypersomething,myhyperparam}. These limitations are compounded by the high runtime of many state-of-the-art AD algorithms, which renders them unsuitable for real-time applications such as user input verification or high-frequency time-series monitoring~\cite{macrodata}.

To address these challenges, we introduce RangeAD, a novel framework that provides highly accurate, application-specific anomaly scores with close-to-zero-shot computational overhead. Our key insight is that the predictive model already contains the information needed to identify anomalies. By exploiting the activation ranges of neurons within the trained model, we can derive a robust indication of "anomalousness" without the need for an external detector. Because these activation ranges are intrinsic to the primary model, they naturally align with the application's specific goals. Crucially, these statistics can be calculated during the standard forward pass of the model, resulting in negligible additional time cost.

Our main contributions are as follows:
\begin{itemize}
\item We introduce the \textbf{On-Model AD} setting, a framework for ML-ready anomaly detection that bridges the gap between theoretical AD and practical deployment.
\item We propose the \textbf{RangeAD} algorithm, which leverages internal neural activation ranges to detect anomalies in real-time.
\item We provide a comprehensive ablation study to validate our design choices and demonstrate the efficacy of our method against current baselines.
\end{itemize}

To facilitate reproducibility and further research, our code is available at \href{https://anonymous.4open.science/r/RangeAD/readme.md}{anonymous.4open.science/r/RangeAD}.

\section{Related Work}

Anomaly Detection (AD) is a long-standing and widely studied problem in machine learning, with a rich body of literature spanning classical statistical methods, modern machine learning approaches, and deep learning-based techniques. Its importance in practical machine learning systems has been repeatedly emphasized in the literature~\cite{positionpaperRoechnerBenchmarking}. Numerous surveys provide comprehensive overviews of the field and its methodological diversity~\cite{survey-ruff,macrodata,surveyzhao,metasurvey}. AD plays a critical role across a wide range of application domains, including fraud detection~\cite{fraudapl,auto_appl_networkintrusion}, industrial and machine fault detection~\cite{faultapl,faultapl2,auto_appl_faultsinparticleaccelators}, and scientific data analysis such as particle physics experiments~\cite{mikuni,auto_appl_particlephysics}. In many real-world systems, anomaly detection operates alongside other machine learning components to ensure reliability, robustness, and safety.

In practice, anomaly detection frequently serves as a monitoring mechanism for machine learning systems deployed in sensitive domains such as healthcare~\cite{OnModelHealth}, Internet-of-Things (IoT) infrastructures~\cite{OnModelIOT}, and financial monitoring systems~\cite{fraudapl}. In these settings, predictive models are typically trained to perform domain-specific tasks (e.g., diagnosis, forecasting, or classification), while anomaly detection is implemented as a separate component tasked with identifying abnormal inputs or system behavior. However, these two systems are commonly developed independently, even though they operate on the same data streams and often rely on related representations. As a consequence, the anomaly detector may fail to fully exploit the information already learned by the primary model, potentially leading to inefficiencies or misalignment between the anomaly detection objective and the model’s operational domain.

A central challenge in anomaly detection is the scarcity or complete absence of labeled anomaly samples. Most AD methods operate in an unsupervised or semi-supervised setting where only normal data is available during training. It is well established that even small amounts of labeled information can significantly improve detection performance by guiding the model toward the most informative subspaces of the data~\cite{labelsMatterMore,survey-ruff,surveyzhao}. Motivated by this observation, we propose the \emph{On-Model AD} setting, which leverages the presence of a supervised model trained for a related task. Although the supervised model itself is not trained for anomaly detection, it encodes task-relevant structure that can provide valuable guidance for identifying abnormal inputs.

This intuition is supported by prior work on neural network representations. Deep neural networks are known to learn hierarchical feature representations, where deeper layers capture increasingly abstract and task-relevant information~\cite{higherlevelfeatures}. Several approaches have exploited this property by applying classical anomaly detection algorithms to the learned representations of neural networks rather than the raw input space~\cite{NNrepresent1,NNrepresent2}. Such approaches can significantly improve the performance of traditional AD algorithms on high-dimensional and complex datasets, as the learned embeddings often provide a more structured representation of the data.

However, representation-based approaches typically suffer from several limitations. Many methods rely on extracting a low-dimensional embedding from the supervised model, which can constrain the information available to the anomaly detector and require architectural modifications or additional training procedures. Furthermore, these approaches often assume strong alignment between the supervised task and the anomalies of interest, which may not hold in practice. They also typically rely on accessing only a small subset of internal representations, limiting the amount of information that can be exploited from the trained model.

Our work takes inspiration from a different property of neural networks: the phenomenon of dead neurons in networks with rectified linear unit (ReLU) activations~\cite{deadneurons}. During training, some neurons become permanently inactive because their inputs never enter the positive activation regime. More broadly, the learned weights and the distribution of training data implicitly constrain the range of activation values that neurons can produce under normal operating conditions, even when using alternative activation functions. This observation suggests that each neuron implicitly defines a feasible activation range determined by the training data and model parameters. Deviations from these learned ranges, therefore, provide a natural signal for detecting anomalous inputs.

Building on this insight, our proposed method leverages activation ranges throughout the network to detect anomalies directly during the forward pass of a trained model. Unlike approaches that train a separate detector or rely on specialized representations, our framework integrates anomaly detection into the predictive model itself. This enables extremely efficient detection with negligible computational overhead while simultaneously leveraging the task-specific knowledge already captured by the pretrained model. As a result, our approach bridges the gap between standalone anomaly detection systems and the predictive models they are meant to safeguard.

\section{Methodology}
\label{sec:rangead}


\subsection{On-Model Anomaly Detection}
\label{sec:rangead_methodology}

We consider the following scenario: We assume a typical machine learning problem, e.g., a classification problem that is trained by applying a neural network to a set of normal data. During deployment, in addition to normal data, anomalies might also appear. Our method then tries to capture these anomalies using both normal data (one-class classification~\cite{deepsvdd}) and the initial neural network's weights. We consider anomaly detection algorithms that can benefit from additional information through a trained neural network as \textbf{On-Model Anomaly Detection}.\\
Given a neural network classifier $f$ with pretrained parameters $\theta_i^k \in \Theta$ for every neuron $n_i^k$ from the layers $l_k \in L$ making up the network $f$ and a dataset of normal samples $x_j \in X_\text{train}$, our method outputs score $s(\hat{x}_j, X_\text{train}, f) \in \mathbb{R}$ that capture the probability of $\hat{x}_j\in X_\text{test}$ being an anomaly. By using a threshold $t$, these scores $s(\hat{x}_j, X_\text{train}, f) > t$ can then be converted to binary decisions with the label $y_\text{ano} \in \{+1, -1\}$ for being an anomaly or not. \\
The proposed methodology comprises three distinct phases: training, preparation, and inference. In the training phase, the initial neural network $f$ is trained on, e.g., a classification problem with labeled data $(x, y) \in X_{\text{train}}$. After that, metrics based on clean data, $M(X_\text{prep},f)$, are extracted from the trained network as preparations. With these, an anomaly score can be assigned to unknown test data during the inference phase. We note that $X_{\text{train}}$ and $X_{\text{prep}}$ can but must not be the same, however they should contain only uncontaminated data, while $X_{\text{test}}$ may contain anomalies.\\

\subsection{RangeAD}

Our instantiation of On-Model AD, which we name RangeAD, requires the following metrics $M(X_\text{prep},f)$: We take the idea that data considered to be normal, e.g., in-distribution, usually stays in a finite, determinable range of values, and everything outside is abnormal. While such a check only finds a limited number of very severe anomalies, the same is true in subspaces of the data, such as those created by the outputs of various neurons in a trained neural network. By considering all neurons of a related neural network, we can thus gain a more decisive anomaly score that also benefits from setting specific idiosyncrasies from the related network.
So for every neuron $n_i^k$ in every observed\footnote{We do not consider every possible layer of the neural network, since e.g. ReLU or Pooling layers only represent degregated copies of previous layers.} layer $l_k$ we calculate the range of normal values in the neural networks feature space as the interval in which clean data produces activation outputs as $I_\text{normal}(n_i^k) = [ Q(n_i^k(X_{\text{prep}}),\sigma), Q(n_i^k(X_{\text{prep}}),1-\sigma)]$ where $n_i^k(X_{\text{prep}})$ is the distribution of observed intermediate activation output of the neuron $n_i^k$ generated in the forward pass of $f(X_{\text{prep}})$ and $Q(\cdot,\sigma)$ represents the $\sigma$-Quantile. While for $\sigma=0$ we consider the maximum observed values, partially contaminated data makes using $\sigma>0$ preferable. 
After the forward pass of test samples $\hat{x} \in X_{\text{test}}$, we check whether the activations lie outside the previously calculated ranges for all of the neurons. The anomaly score ($s(\hat{x})$) for a sample $\hat{x}$ is then the cardinality of activations lying outside the respective intervals.
\begin{equation}
    s(\hat{x}) = \sum_{k}^{L} \sum_i^{N_k} \Theta(n_i^k(\hat{x}) \notin I_\text{normal}(n_i^k))\hspace{2em} \Theta(\text{True})=1, \Theta(\text{False})=0
\end{equation}
where $\Theta(\cdot)$ is a function that returns 1 if the statement is true and 0 otherwise.
A higher count of out-of-range activations indicates a greater likelihood that the sample is an anomaly. Figure \ref{fig:explain_fig} illustrates this process.
\begin{figure}
\centering
\begin{subfigure}{0.25\textwidth}
    \centering
    \begin{tikzpicture}[
        scale=1.0,
        neuron/.style={circle, draw=none, minimum size=4mm},
        blue/.style={neuron, fill=blue!70},
        orange/.style={neuron, fill=orange!80},
        green/.style={neuron, fill=green!70},
        connect/.style={black!60, thin}
    ]
        \def\layercenter{1.2}      
        \def\layerspacing{0.8}    
        
        \def\horizontalspacing{1.3}
        
        \node at (0,3.4) {$l_1$};
        \node at (\horizontalspacing,3.4) {$l_2$};
        \node at (1.5*\horizontalspacing,3.4) {...};
        \node at (2*\horizontalspacing,3.4) {$l_N$};
        
\foreach \i in {1,2,3} {
    \node[blue] (l1\i) at (0,{\layercenter + (2-\i)*\layerspacing}) {};
}

\foreach \i in {1,2,3,4} {
    \node[orange] (l2\i) at (\horizontalspacing,{\layercenter + (2.5-\i)*\layerspacing}) {};
}

\foreach \i in {1,2} {
    \node[green] (l3\i) at (2*\horizontalspacing,{\layercenter + (1.5-\i)*\layerspacing}) {};
}
        
        \foreach \i in {1,2,3} {
            \foreach \j in {1,2,3,4} {
                \draw[connect] (l1\i) -- (l2\j);
            }
        }
        \foreach \i in {1,2,3,4} {
            \foreach \j in {1,2} {
                \draw[connect] (l2\i) -- (l3\j);
            }
        }
    \end{tikzpicture}
    \caption{}
    \label{fig:expl_fig_network}
\end{subfigure}
\hfill
\begin{subfigure}{0.35\textwidth}
    \centering
    \begin{tikzpicture}[scale=0.6]
        \begin{scope}
        \pgfmathsetseed{124}
        \foreach \y in {2.5,1.5,0.5} {
            \pgfmathsetmacro{\xmin}{1000}
            \pgfmathsetmacro{\xmax}{-1000}
            
            \foreach \k in {1,...,12} {
                \pgfmathsetmacro{\rx}{rnd*3.4}
                \pgfmathsetmacro{\ry}{\y + (rnd-0.5)*0.1}
        
                \pgfmathparse{min(\xmin,\rx)}\xdef\xmin{\pgfmathresult}
                \pgfmathparse{max(\xmax,\rx)}\xdef\xmax{\pgfmathresult}
                
                \fill[blue!70] (\rx,\ry) circle (1.6pt);
            }
            
            \draw[blue!70, thick] (\xmin,\y-0.35) -- (\xmax,\y-0.35);
            \draw[blue!70, thick] (\xmin,\y-0.4) -- (\xmin,\y-0.3);
            \draw[blue!70, thick] (\xmax,\y-0.4) -- (\xmax,\y-0.3);
        }
        \node at (1.7, -0.5) {$I_\text{normal}(n_i^1)$};
        
        \foreach \y in {3.5,2.5,1.5,0.5} {
        
            \pgfmathsetmacro{\xmin}{1000}
            \pgfmathsetmacro{\xmax}{-1000}
            
            \foreach \k in {1,...,12} {
                \pgfmathsetmacro{\rx}{4.0 + rnd*3.4}
                \pgfmathsetmacro{\ry}{\y + (rnd-0.5)*0.1}
        
                \pgfmathparse{min(\xmin,\rx)}\xdef\xmin{\pgfmathresult}
                \pgfmathparse{max(\xmax,\rx)}\xdef\xmax{\pgfmathresult}
                
                \fill[orange!70] (\rx,\ry) circle (1.6pt);
            }
            
            \draw[orange!70, thick] (\xmin,\y-0.35) -- (\xmax,\y-0.35);
            \draw[orange!70, thick] (\xmin,\y-0.4) -- (\xmin,\y-0.3);
            \draw[orange!70, thick] (\xmax,\y-0.4) -- (\xmax,\y-0.3);
        }
        \node at (5.5, -0.5) {$I_\text{normal}(n_i^2)$};
        \end{scope}
    \end{tikzpicture}
    \caption{}
\end{subfigure}
\hfill
\begin{subfigure}{0.25\textwidth}
    \centering
    \begin{tabular}{c c c}
        \begin{tabular}{@{}c@{}}Check intervals\end{tabular} & \\
        & \\
        ${n_1^{1}}(x_i) \in I_\text{normal}(n_1^1)$ \\ [1.5ex] 
        ${n_2^{1}}(x_i) \in I_\text{normal}(n_2^1)$ \\ [1.5ex] 
        ${n_3^{1}}(x_i) \in I_\text{normal}(n_3^1)$ \\ [1.5ex] 
        ... & \\
    \end{tabular}
    \caption{}
\end{subfigure}
\caption{\textbf{Depiction of our proposed methodology}. (a) represents the neural network, from which the sample activations per neuron are depicted in (b). From these distributions - to be considered \textit{normal} - interval borders are derived. The activations produced for a new test input are then checked per neuron to see whether it falls inside the corresponding interval in (c). For every test sample $x_i$ \textbf{the amount of feature-outputs lying outside of the corresponding interval indicates the anomaly grade} of $x_i$.}
\label{fig:explain_fig}
\end{figure}\\
To convert continuous anomaly scores into binary decisions, the default approach selects a threshold based on the expected number of anomalies in the test set~\cite{pyod}. When using RangeAD, we can also select a more intuitive threshold based on the fraction of observed neurons (e.g. $10\%$).

\section{Evaluation}
\label{sec:evaluation}

In this section, we present a comprehensive empirical evaluation of our proposed anomaly detection framework. We benchmark our method against several baselines across three distinct data modalities. For each experimental scenario, out-of-distribution (OOD) test sets are systematically curated by either introducing external anomalous samples or selectively filtering existing classes to ensure contextually relevant detection challenges. In every experiment, our framework is built on a classification neural network trained on clean data (in-distribution) in the respective domain, and the detection performance is evaluated on a contaminated testing dataset. Our primary experiments focus on tabular data, which remains a cornerstone of anomaly detection research. To demonstrate the versatility of our approach, we extend our evaluation to image and time-series classification tasks. 


Finally, we conduct a detailed ablation study to assess the sensitivity of our method to various implementation choices and hyperparameters.\\
For the vision and time series tasks, we took $\sigma=1\% / 0.1\%$ quantiles to derive the interval borders from the activation distribution and the $\sigma=1\%$ quantile for the tabular data, for which other variants are covered in the corresponding ablation study. 
To measure the efficiency of our algorithm, we separate the required runtime into three phases. Both the training and forward pass of the initial neural network occur whether we use it for on-model anomaly detection or not, and are thus ignored here. The training time of our method (computing $M(X_\text{prep},f)$) occurs only once and is thus usually negligible in practice. Hence, we focus here on the inference time, as this represents the time required to classify new samples and state the remaining times in the supplementary material. All experiments were performed using an Intel Xeon Gold 6258R CPU, 252 GB RAM, and a Nvidia Quadro RTX 6000 GPU with 24 GB VRAM.

\subsection{Anomaly detection performance}
\label{sec:eval_ad_performance}

\subsubsection{Tabular data}
\begin{figure}
    \centering
    \includegraphics[width=0.99\linewidth]{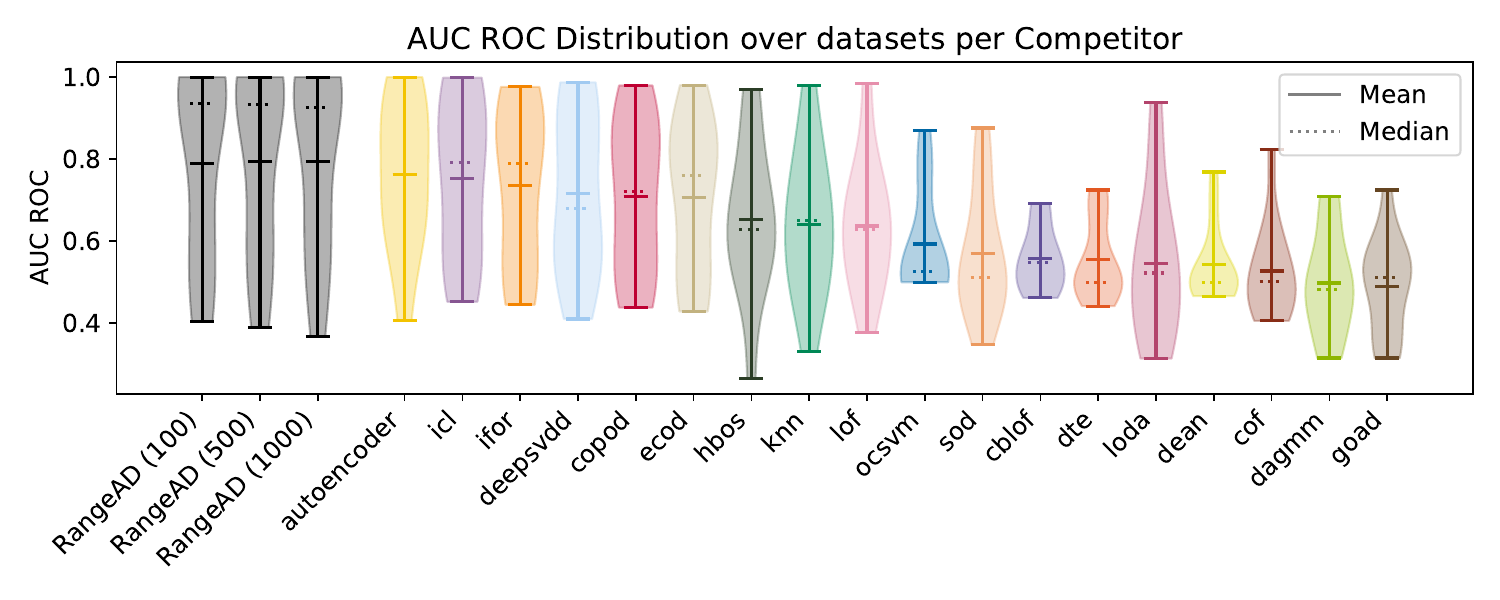}
    \caption{AUC-ROC of different anomaly detection methods over 12 tabular datasets. Our method is run three times with models having hidden-layer neuron counts of 100, 500, and 1000. \textbf{It reaches the highest AUC-ROC on average compared to all competitors.}}
    \label{fig:roc_auc_tabular}
\vspace{-1em}
\end{figure}
The 12 tabular datasets are selected from OddBench~\cite{macrodata} by taking all datasets with multiple normal classes and at least $30$ features, for which our networks achieve at least 70\% classification accuracy on the clean part of the test data~\footnote{We filter for the performance of the related network, as we desire models that can be deployed in practice.}. Thus, we take the "BaseballEvents", "CasePriority", "CreativeSchoolCertification", "FinanceJobCategories", "GamePositionAnomaly", "HousingVulnerability", "NFLPerformanceAnomalies", "SentencingFraud", "TravelWeatherScores", "TreeCondition", "WeatherUVIndex", and "WindowsEditionAnomaly" datasets, which each contain multiple normal classes in addition to anomalies. For our experiments, we train a simple three-layer Multilayer Perceptron (MLP) with ReLU as a non-linear activation function on the normalized, multi-class training data as a classifier. We employ hidden layers with 100, 500, and 1000 neurons to achieve diversity, roughly 2-50 times the dataset's feature count. We take the outputs of the first two hidden dimensions (before ReLU activation) for calculating the anomaly detection ranges. In the supplementary material, we compare alternative layer choices for our method. The neural networks are trained for 500 epochs with a learning rate of 0.001. While evaluating our anomaly detection method, we use the training dataset during the building phase and add the anomalous class to the test data for evaluation.\\
We evaluate the proposed method mainly on runtime and the Area Under the Receiver Operating Characteristic Curve (AUC-ROC) and compare it with several other anomaly detection techniques from the literature on these two metrics across the same 12 tabular datasets. The competitors feature Autoencoder~\cite{aean}, CBLOF~\cite{cblof}, COF~\cite{cof}, COPOD~\cite{copod}, Dagmm~\cite{dagmm}, Dean~\cite{DEAN}, DeepSVDD~\cite{deepsvdd}, DTE~\cite{dte}, ECOD~\cite{ecod}, GOAD~\cite{goad}, HBos~\cite{hbos}, ICL~\cite{icl}, Isolation Forest~\cite{ifor}, k-Nearest Neighbour~\cite{knn}, LODA~\cite{loda}, LOF~\cite{lof}, OC-SVM~\cite{ocsvm} and SOD~\cite{sod} providing various angles on anomaly detection. \\
The anomaly detection results can be seen in Figure \ref{fig:roc_auc_tabular}. Our method achieves the highest AUC-ROC most often on the datasets compared to the competitors. Additionally, all three model sizes lead to an average AUC-ROC higher than all competitors by a difference of 0.03 to the second-best approach (0.79 for our worst model compared to 0.76 for the autoencoder). In seven of the twelve datasets, the top AUC-ROCs were above 0.9, while our approach most often ranked among the top three competitors. In the remaining five datasets, our method achieved AUC-ROCs still in the better half of the methods compared. Having the highest average score across all competitors and all three model sizes shows that our approach is competitive with all the compared methods. We provide a critical difference plot in the supplementary material.\\

Figure \ref{fig:roc_auc_vs_runtime_tabular} shows the results of the runtime evaluation. Our method has an average inference time of around 2 ms across all three model sizes, while the second-best competitor, an Autoencoder, is roughly 100 times slower. The fastest competitor, DeepSVDD, still needs almost double the runtime for an average inference pass over the test dataset, 3.7 ms. The results highlight the benefits of the on-model AD setting, as our method is the fastest and best-performing algorithm among those used. It outperformes the deep and slower methods detection wise and at the same time is faster than the lightweight competitors.\\
Comparing the three hidden layer sizes, used in our method, the wider networks seem to perform better while naturally being a bit slower. This indicates a tradeoff, which we will cover in more detail in the ablation study.\\
\begin{figure}
    \centering
    \includegraphics[width=0.99\linewidth]{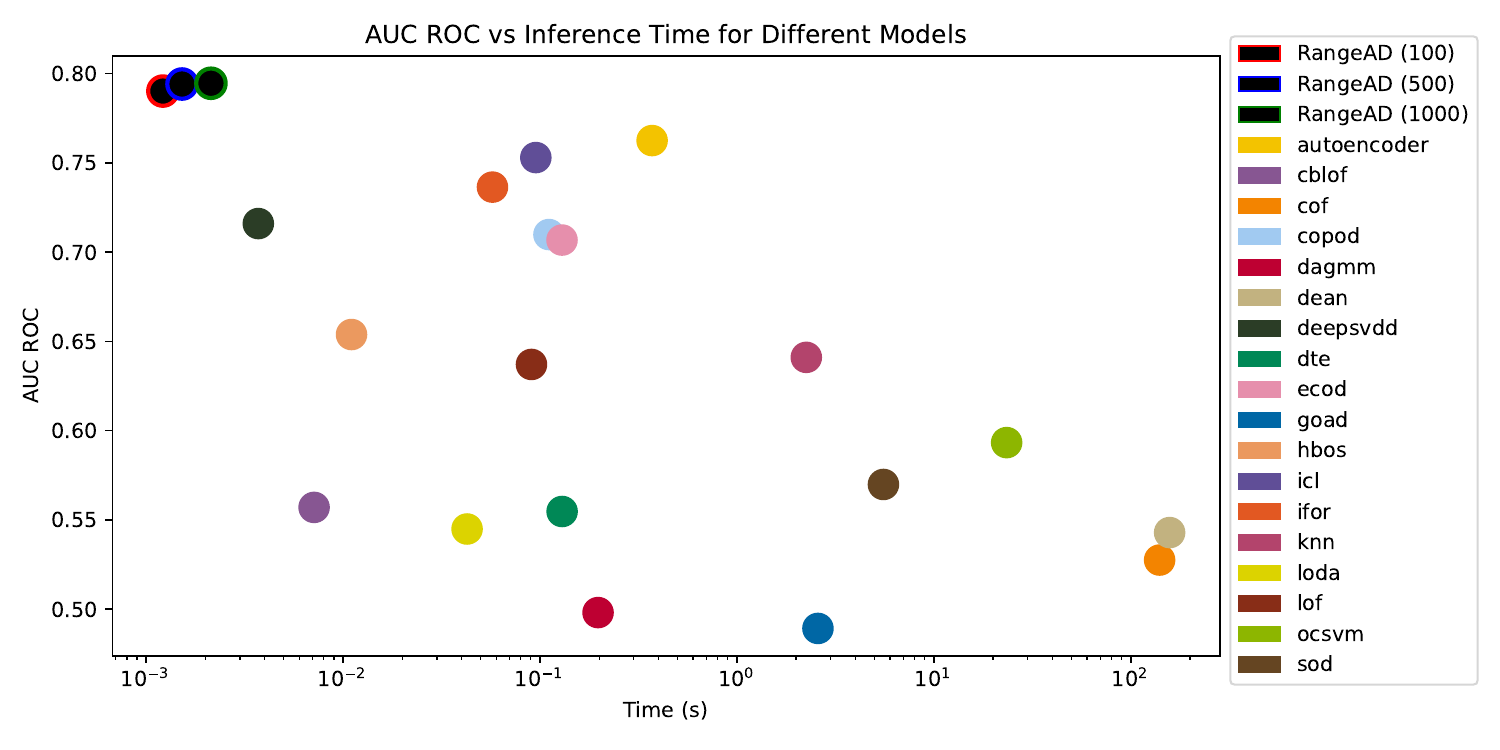}
    \caption{AUC-ROC and inference time of different anomaly detection methods averaged over 12 tabular datasets. Our method is run three times with models with hidden layer neuron counts of 100, 500, 1000. \textbf{It achieves the highest detection performance while providing the fastest inference runtime of all competitors.} We provide the same analysis on the training time in the supplementary material.}
    \label{fig:roc_auc_vs_runtime_tabular}
    \vspace{-3em}
\end{figure}

\subsubsection{Vision Data}
While the previous section shows that our method achieves comparable or even better performance to the competitors on relatively low-dimensional tabular data, we also aimed to evaluate on more complex scenarios. Therefore, we also conduct anomaly detection experiments on image data. Therefore, we take a modern Swin-v2-Vision Transformer~\cite{liu2022swintransformerv2scaling} model pretrained on ImageNet-1k (in-distribution) as our base network. The feature spaces from which the detection intervals are calculated are either taken after every GELU-Activation function call ("in") or after a whole Transfomer-Block ("after"). We note that all neurons could also be used at the cost of a higher memory usage, but for our experiments, the 17664/4416 neurons (GELU/Transformer-Block) were already enough to perform well. We utilize the 100,000 ImageNet test images as the preparation dataset for building our anomaly detector. The test dataset to evaluate on consists of the whole Imagenette subset~\cite{imagenette} (train+test) as in-distribution and two different datasets from domains that are not contained in ImageNet-1k, as out-of-distribution, namely AstronomyImages~\cite{spaceimages} \footnote{Classes "cosmos space", "galaxies", "nebula" and "stars" were used as they provide the most similar photographs and ImageNet-1k does only contain very few space based images.} and Ambivision~\cite{ambivision}, a dataset of animal-based optical illusions, to bring in multiple angles for the OOD-domain. We perform the evaluation on the two OOD-datasets, both separately and combined, and compare our method against an Autoencoder, Isolation Forest, and DeepSVDD, the top 3 competitors from the tabular experiment, excluding ICL for computational reasons.
\begin{table}
    \centering
\resizebox{\textwidth}{!}{%
    \begin{tabular}{cccccccc}\toprule
 Dataset& in, 0.01& after, 0.01& in, 0.001&after, 0.001& IF&AE &DeepSVDD\\\midrule
         INet + Ambivision&  0.9893&  0.9923&   0.9945&0.9937& \textbf{0.9985}& 0.9472&0.8830\\
         INet + SpaceImages&  \textbf{0.9625}&  0.9094&   0.9592&0.909& 0.4977& 0.1864&0.6757\\
         INet + both&  \textbf{0.971}&  0.9355&   0.9703&0.9357& 0.6597& 0.4266&0.6833\\\midrule
         UrbanSound~\cite{salamon2014dataset}&  0.8804&  -&   \textbf{0.9142}&-& 0.6939 & 0.8002&0.7927\\
         \bottomrule 
    \end{tabular}
    }
    \caption{\textbf{AUC-ROC} of vision and time series tasks. "in" means in the network blocks, e.g. after GELU and Convolution-Layers; "after" means after the network blocks, e.g. Transformer-Block (vision-only). 0.01 and 0.001 is the quantile used. AE and DeepSVDD were trained with 10\% of the INet testset for memory reasons. \textbf{Our method reaches the most reliable performance across high-dimensional datasets.}}
    \label{tab:exp_rocauc_ts_vis}
    \vspace{-2em}
\end{table}\\
The results from this experiment show that also in the vision anomaly detection task, our approach can compete with the baselines and even outperforms them in almost every setup. In Table \ref{tab:exp_rocauc_ts_vis} the AUC ROCs can be found. When taking Ambivision as the OOD-dataset, the Isolation Forest has an AUC ROC negligibly (0.004) higher than the best variant of our method, when taking activations after GELU output and using 0.1\% quantile. Despite that, with the SpaceImage as OOD our using both OOD-Datasets, all variants of our method perform atleast 0.2 AUC ROC better than the competitors. Internally, the GELU-feature space tends to provide a better anomaly detection capability than the outputs at the end of the transformer-block. We discuss alternative layer choices in the supplementary material. 
The inference runtimes provide even better results of our method compared to the three competitors. Table \ref{tab:exp_time_ts_vis} shows that our method takes roughly 0.014 seconds for performing inference. The Isolation Forest performs inference in roughly 0.8 seconds, over 50 times more than ours, while the two deep approaches take several minutes. Across the different variants tested for our method, there is almost no difference in runtime, whether the observed layers or the quantiles are varied, highlighting the scalability of RangeAD.

\begin{table}
    \centering
\resizebox{\textwidth}{!}{%
    \begin{tabular}{cccccccc}\toprule
 Dataset& in, 0.01& after, 0.01& in, 0.001&after, 0.001& IF&AE&DeepSVDD\\\midrule
         Inet + Ambivision&  0.0151&   0.0161&   0.0131&\textbf{0.0129}& 0.8197&240.9355&133.9441\\
         Inet + SpaceImages&  0.0131&  \textbf{0.012}&   0.0123&0.0129& 0.8438&238.3285&138.8146\\
         Inet + both&  0.0129&  \textbf{0.0126}&   0.0129&0.0136& 0.8560&251.041&141.8705\\\midrule
         UrbanSound~\cite{salamon2014dataset}& 0.0138&  -&  \textbf{0.0103}&
 -&0.0711& 0.8833&0.5184\\
         \bottomrule 
    \end{tabular}
    }
    
    \caption{\textbf{Inference time} in seconds of vision and time series tasks. "in" means in the network blocks, e.g. after GELU and Convolution-Layers; "after" means after the network blocks, e.g. Transformer-Block (vision-only). 0.01 and 0.001 are the quantiles used. AE and DeepSVDD were trained with 10\% of the INet testset for memory reasons. \textbf{Our method requires multiple orders of magnitude less runtime as it is not using an additional model.}}
    \label{tab:exp_time_ts_vis}
    \vspace{-4em}
\end{table}

\subsubsection{Time Series}

In addition to tabular and image data, we wanted to evaluate our method in a third domain to demonstrate its general applicability. We choose time-series data and use the UrbanSound dataset~\cite{salamon2014dataset}, which consists of 44.1kHz recordings of typical urban environments, for acoustic analysis. The dataset consists of ten classes representing common sounds from the urban area like children playing, dogs barking or construction site noise. Within this framework, gunshot sounds are designated as the OOD class, as they represent sparse, high-impact events that are naturally anomalous to the urban soundscape. This scenario is modeled using a temporal convolutional network (TCN)~\cite{tcn} coupled with an MLP classification head for classifying the sound data. We train the model for 20 Epochs on the normal classes achieving roughly 65\% test accuracy. To calculate the anomaly detection ranges, we use the output of the convolution layers. After building our anomaly detection method, we add the anomaly class to the test data for evaluating. We compare the results to the same three competitors from the vision task, Autoencoder, Isolation Forest and DeepSVDD.\\
Table \ref{tab:exp_rocauc_ts_vis} and Table \ref{tab:exp_time_ts_vis} show the AUC ROCs and inference times respectively. While the best competitor - an Autoencoder - achieves around 0.8 AUC ROC, our method detects anomalies with an AUC ROC around 0.9. The difference between using the 1\% and 0.1\% quantile is over 0.03, comparable to the difference in the vision tasks at around 0.01, highlighting that tuning this hyperparameter can slightly increase the performance depending on the scenario, but this is not deeply necessary. The inference runtime is with 0.013 seconds on par with the vision task, as it is only dependent on the sample count and the neurons used, not the feature size, which would be the most influential difference to the time series dataset. The three competitors, on the other hand, scale worse with higher feature dimensions, needing from 0.07 seconds to 0.88 seconds, showing that our method is also more efficient and effective for time-series anomaly detection.

\subsection{Ablation Study}
\label{sec:eval_abl}

To understand the details and alterations of our method, we investigate different aspects of the approach in an ablation study. Therefore we evaluate the behavior in different scenarios on the tabular datasets.

\subsubsection{Model Size Ablations}

\begin{figure}[!t]
    \centering
    \includegraphics[width=0.95\linewidth]{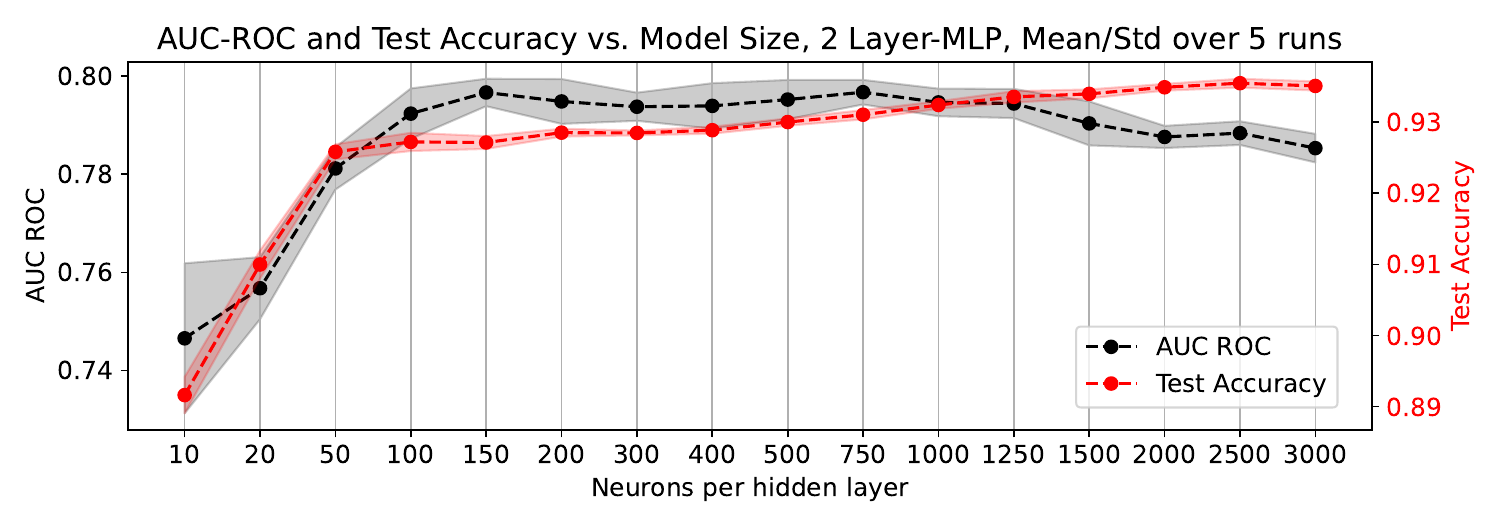}
    \caption{When using differently sized hidden layers in our neural networks, \textbf{AUC ROC and model performance are correlated}. The results represent the average of 5 runs with a standard deviation over the 12 tabular datasets.}
    \label{fig:roc_auc_test_acc_vs_model_size}
\end{figure}

First, we want to investigate how model size and the feature space size affect anomaly detection performance. For this, we alter the tabular setup so that the initial classifier consists of hidden layers with 10 to 3000 neurons. By taking the feature space of the first two layers of the MLP-Network, our method detects anomalies based on 20 to 6000 activation intervals. For each model size, the detection performance was captured as the AUC ROC.\\
While the accuracy of the initial classifier increases logarithmically and interestingly does not observe overfitting, as shown in Figure \ref{fig:roc_auc_test_acc_vs_model_size}, the detection performance mostly closely follows this pattern. They both exhibit sharp initial gains at smaller scales, followed by shallower increases at larger layer sizes. As test accuracy continues to improve with more than 1000 neurons per layer, the AUC ROC reaches its peak at 750 neurons and then declines slightly, indicating modest oversaturation.\\
\subsubsection{Model Training Ablations}

\begin{figure}
\centering
\begin{subfigure}{0.49\textwidth}
    \includegraphics[width=\textwidth]{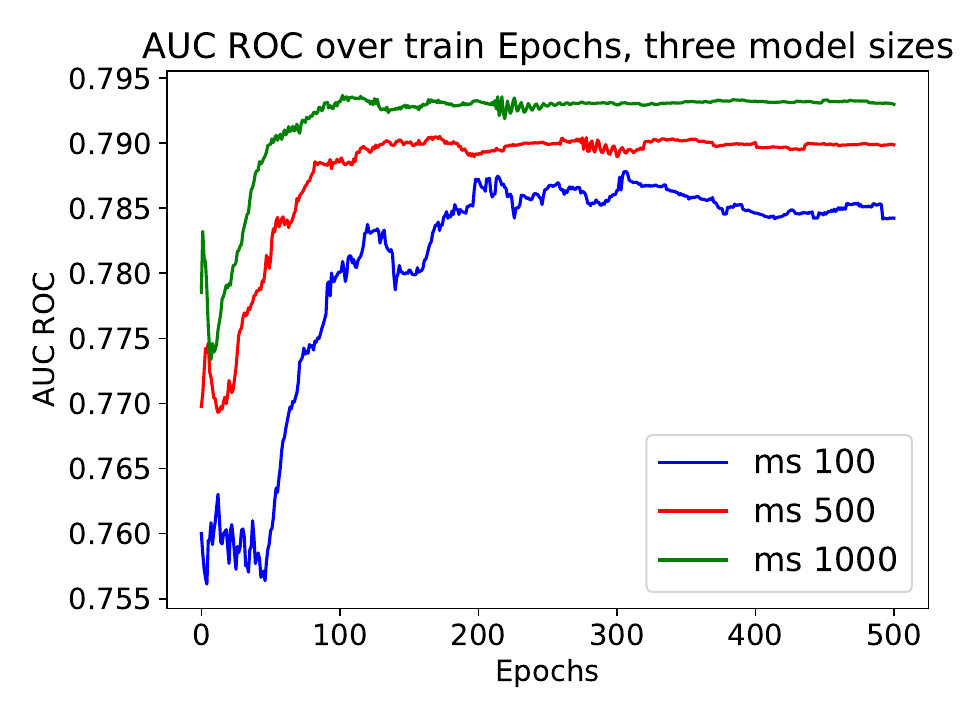}
    \caption{Anomaly detection performance as AUC-ROC of our method in three different model sizes measured after every training epoch of the initial classifier. \textbf{Even with minimal training our method reaches competitive performance.}}
    \label{fig:roc_auc_vs_epochs_all_model_sizes}
\end{subfigure}
\hfill
\begin{subfigure}{0.49\textwidth}
    \includegraphics[width=\textwidth]{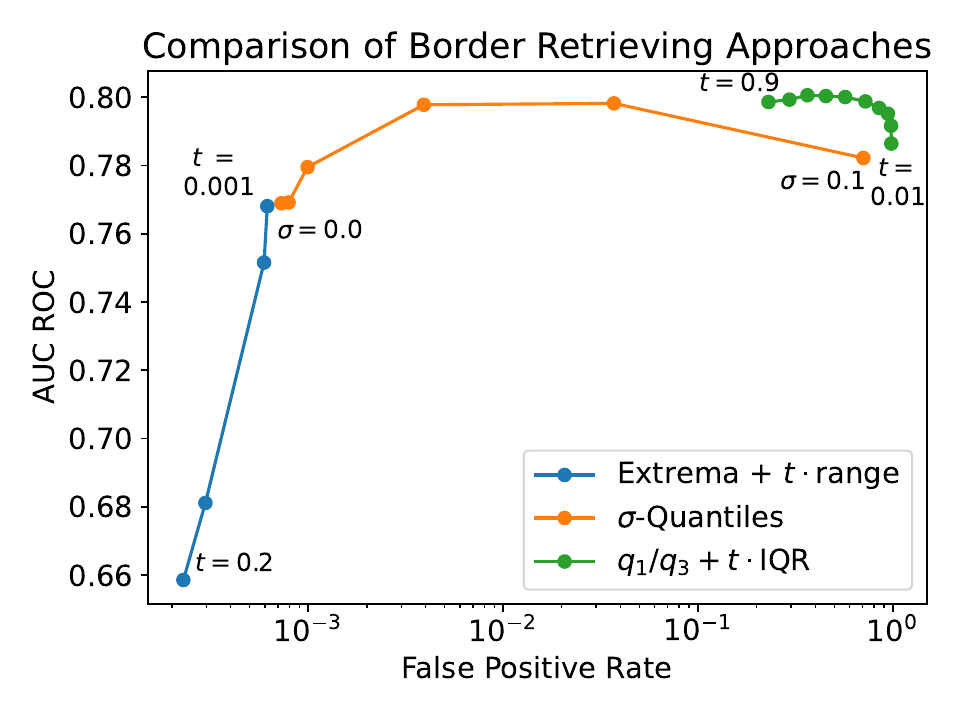}
    \caption{AUC ROC and False-Positive-Rate (FPR) for different approaches of calculating the interval borders. \textbf{The Quantile-based approach reaches the best balance between FPR and anomaly detection performance.}}
    \label{fig:border_retrieval_distance}
\end{subfigure}  
\caption{Ablation study results.}
\label{fig:ablation_study_figures}
\end{figure}

After investigating the model size, we seek to understand the influence of the training of the base network. Our assumption here is that with a better-performing / longer-trained network, the activation ranges become more meaningful, leading to a more precise anomaly separation and a superior detection performance. To achieve this comparison, we train the initial classifier and measure detection performance after every epoch. As the previous experiment showed a dependency of the performance on different model sizes and to provide more meaningful outcomes, we use 100, 500, and 1000 neurons in the MLP's hidden layers.\\
Going from the results in Figure \ref{fig:roc_auc_vs_epochs_all_model_sizes}, we note that across the three tested model sizes, the detection performance went higher over the epochs, indicating a generally better performance with better-trained models. Additionally, we note that even before training, the AUC-ROC was high. Compared with the competitors in Figure \ref{fig:roc_auc_tabular}, our models achieve an average ROC AUC higher than almost all competitors, even in an untrained state. From this, we relate that the intermediate feature spaces of neural networks are intrinsically good indicators for anomalies. By training these spaces, one can further improve these capabilities, but our method also likely still works very well when the related model is unrelated to the type of anomalies we are searching for.

\subsubsection{Range Border Retrieval}

The final ablation study evaluates different methods for determining the borders of the \textit{normal}-intervals derived from each neuron's activation output. Our primary approach uses quantiles close to the extrema of the activation distribution. We therefore evaluate different values of $\sigma$, which control the distance of the selected quantiles from the distribution boundaries, in a tabular setting using an MLP with 1000 hidden units. We additionally consider approaches that place borders further from the extrema. First, we use the quartiles $(q_1, q_3)$ and extend them by adding a fraction $t$ of the interquartile range (IQR), producing borders between the quartiles and the extrema. Second, we apply the same idea using the minimum and maximum together with the Min-Max range, resulting in borders outside the observed activation distribution. The scaling parameter $t$ varies from 0.001 to 0.9, while $\sigma$ ranges from 0 to 0.1.\\
Borders placed deep within the distribution classify many samples as anomalies, whereas borders outside the distribution yield few detections. To analyze this trade-off, we compare the AUC-ROC to the resulting false-positive rate (FPR). Binary predictions for the FPR are obtained using a threshold based on 10\% of the monitored neurons (200 of 2000 for two layers).
Figure \ref{fig:border_retrieval_distance} shows the resulting AUC-ROC-FPR trade-offs. The extrema-based method yields the lowest FPR but also the lowest AUC-ROC (maximum at 0.76). The quantile- and quartile-based approaches achieve similar AUC-ROC values close to 0.8, although the quartile-based method produces substantially higher FPR. Overall, the quantile-based approach provides the most favorable trade-off in this setting. 

\section{Conclusion and Future Work}
\label{sec:conclusion}

In this paper, we introduce a novel anomaly detection setting that exploits a related machine learning model, which we call On-Model AD. We also introduce a technique that can leverage such a setting, RangeAD. Our scenario and algorithm are especially useful in complicated production environments, where a related neural network model often exists to solve task beyond anomaly detection, e.g., classification.
By leveraging the neural network's intermediate feature space and analyzing the activation outputs of the preparation data, our method constructs a single interval of \textit{normal} activations per neuron. Based on this building phase, our method detects anomalies by aggregating the information wether the observed data's activation falls inside or outside each neurons interval.\\
We examined our method in anomaly detection scenarios across tabular, vision, and time series domains, in each of which it proved a superior performance to the competitors, doing so with regards to both detection quality and efficiency. In an ablation study, we further analyzed the approach and made multiple changes to the detection framework to highlight its variations and limitations.\\
While this study focused on point-wise anomalies within static classification tasks, the framework’s versatility suggests high extensibility to regression and streaming time-series environments. A particularly compelling frontier lies in Natural Language Processing. Given the scalability demands and the problem of anomalous Large Language Models states (hallucinations)~\cite{hallucinationAsAnomalies},
our lightweight add-on approach could enhance the reliability in validating LLM outputs. Future research will explore architectural optimizations to further refine this on-model detection, including selective neuron pruning, automated optimal layer selection, and specialized regularization techniques designed to cultivate more discriminative feature representations during training. These advancements aim to further reduce computational overhead while maximizing the sensitivity of the anomaly detection boundary.\\



%
%
%

\subsubsection{\ackname}
This research was supported by the Research Center Trustworthy Data Science and Security (\url{https://rc-trust.ai}), one of the Research Alliance centers within the University Alliance Ruhr (\url{https://uaruhr.de}).

\bibliographystyle{splncs04}
\bibliography{simons_refs, references}

\end{document}